\documentclass[11pt]{article}

\usepackage{amsmath}
\usepackage{amssymb}
 \usepackage{booktabs}
\usepackage[preprint]{acl}

\usepackage{times}
\usepackage{latexsym}

\usepackage[T1]{fontenc}

\usepackage[utf8]{inputenc}

\usepackage{microtype}

\usepackage{inconsolata}

\usepackage{graphicx}
\usepackage{multirow}
\usepackage{float}
\newcommand{\state}[1]{\text{\textsc{#1}}}
\title{Weighted Memory Tree: Remembering What Matters for Long-Horizon LLM Agents}

\author{
\textbf{Quang Dao}\textsuperscript{1},
\textbf{Purvi Kathalkar}\textsuperscript{2},
\textbf{Kenneth Eaton}\textsuperscript{3} \\
\textsuperscript{1}Rose-Hulman Institute of Technology \\
\textsuperscript{2}Georgia Institute of Technology,
\textsuperscript{3}Georgia Tech Research Institute
}

\begin{document}
\maketitle
\begin{abstract}
Large language model (LLM) agents have demonstrated the ability to solve multi-step tasks requiring planning, tool use, and external information access, yet growing execution histories increase inference cost and expose reasoning to outdated, irrelevant, or misleading information, potentially degrading reasoning quality.
Existing memory approaches organize or compress execution histories but provide limited mechanisms for deciding which memories remain active. 
We introduce the \textbf{Weighted Memory Tree} (\textbf{WMT}), a hierarchical memory system that organizes execution into tasks, subtasks, and actions while assigning each memory a dynamic retention score. 
Event-based updates and selection-based decay revise these scores, allowing WMT to preserve useful information, fold completed trajectories, suppress low-utility content, and retain access to folded context. 
We evaluate WMT on GAIA-Text using Qwen3-8B, Gemma 4 E4B, and
Llama-3.1-8B, with ablations and memory-poisoning experiments.
Relative to linear memory, WMT improves accuracy by an average of
9.97 percentage points while reducing prompt-token usage by 32.8\%.
Memory-poisoning experiments show that WMT limits the persistence and propagation of unreliable information. 
Our results suggest that effective long-horizon agent memory depends less on storing more information than on deciding which information should remain active.
\end{abstract}

\section{Introduction}
Large language model (LLM) agents interleave reasoning with actions in external environments, enabling them to retrieve information, invoke tools, and revise plans over multiple steps to solve long-horizon tasks such as open-domain research \citep{yao2023react,schick2023toolformer,sun2026scaling}. 
Such tasks require agents to preserve evidence, tool outputs, and failed attempts encountered earlier in the execution, making memory central to maintaining the task state required for subsequent reasoning \citep{zhang2025memorysurvey,guan2026evaluating}. 
As execution histories grow, however, outdated observations, failed attempts, and incidental details accumulate alongside useful information, making it increasingly difficult to determine which memories should continue to influence subsequent decisions.

A common ReAct-style design preserves execution state by appending reasoning steps, tool calls, and observations to a linear interaction history \citep{yao2023react}. 
Although this retains the full trajectory, it treats all memories equally, regardless of their utility, causing prompt lengths to grow while outdated observations, failed reasoning, and valid evidence remain intermixed. 
Consequently, relevant information must compete with stale or incidental content, a limitation that is not resolved simply by expanding the context window.
Long-context studies have shown that language models struggle to utilize relevant information within lengthy inputs, with performance often degrading well below the nominal context-window limit \citep{liu2024lost,levy-etal-2024-task,fraga2024challenging,he-etal-2024-never,hsieh-etal-2024-found,tian-etal-2025-distance}. 
Beyond efficiency and reasoning quality, persistent agent memory also introduces a security concern: poisoned records retrieved from long-term memory or external knowledge bases can influence subsequent agent behavior \citep{chen2024agentpoison, srivastava2025memory, FERRAG2026353, dong2025memory}. 
Long-horizon agents, therefore, require mechanisms that regulate which stored information enters the working context.

Recent work addresses parts of this problem through structured or compressed representations of execution history. 
The Task Memory Engine (TME) organizes execution as a hierarchical task tree and synthesizes prompts from the active task path \citep{ye2025taskmemory}. 
Context-Folding summarizes completed subtasks before returning to the parent trajectory \citep{sun2026scaling}. 
These approaches establish the value of preserving task structure and compressing completed trajectories. 
However, they do not explicitly maintain a dynamic estimate of each memory’s continuing utility. As a result,
memories may remain active even after subsequent evidence reduces their relevance or reliability. 
This motivates a central question: can long-horizon reasoning be improved by explicitly estimating memory utility to regulate what enters the agent’s working context?

To address this challenge, we introduce the \textbf{Weighted Memory Tree}
(\textbf{WMT}), a hierarchical memory architecture that organizes
execution histories into task, subtask, and action memories while assigning
each memory a dynamic retention score. 
These scores prioritize memories for context construction and determine
when low-utility branches are suppressed, while completed branches are
folded into compact summaries according to task status, allowing WMT to construct prompts from the most useful context while reducing repeated token processing and limiting the influence of stale or unreliable information \citep{chen2024agentpoison}.
We evaluate WMT on GAIA \citep{mialon2024gaia} and its text-only subset, GAIA-Text, using Qwen3-8B \citep{yang2025qwen3}, Gemma 4 E4B \citep{gemma4}, and Llama-3.1-8B \citep{grattafiori2024llama3}. 
We compare the full framework against a linear-memory baseline and three component ablations: an unweighted tree (A1), a no-memory controller ablation (A2), and an ablation without semantic node retrieval (A3).

Although mitigating memory poisoning is not the primary objective of WMT, we also conduct controlled memory-poisoning experiments to evaluate whether structured memory management improves robustness relative to conventional linear memory.
Across these experiments, WMT reduces attack success rate, poison retrieval rate, blast radius, and amplification factor while achieving the highest task success rate among all evaluated methods. 
Together, these results suggest that effective long-horizon memory depends not only on preserving or compressing execution history, but on selectively retaining task-relevant information while suppressing outdated or unreliable content.

\begin{figure}[!t]
  \centering
  \includegraphics[width=\columnwidth, height=0.35\textheight, keepaspectratio]{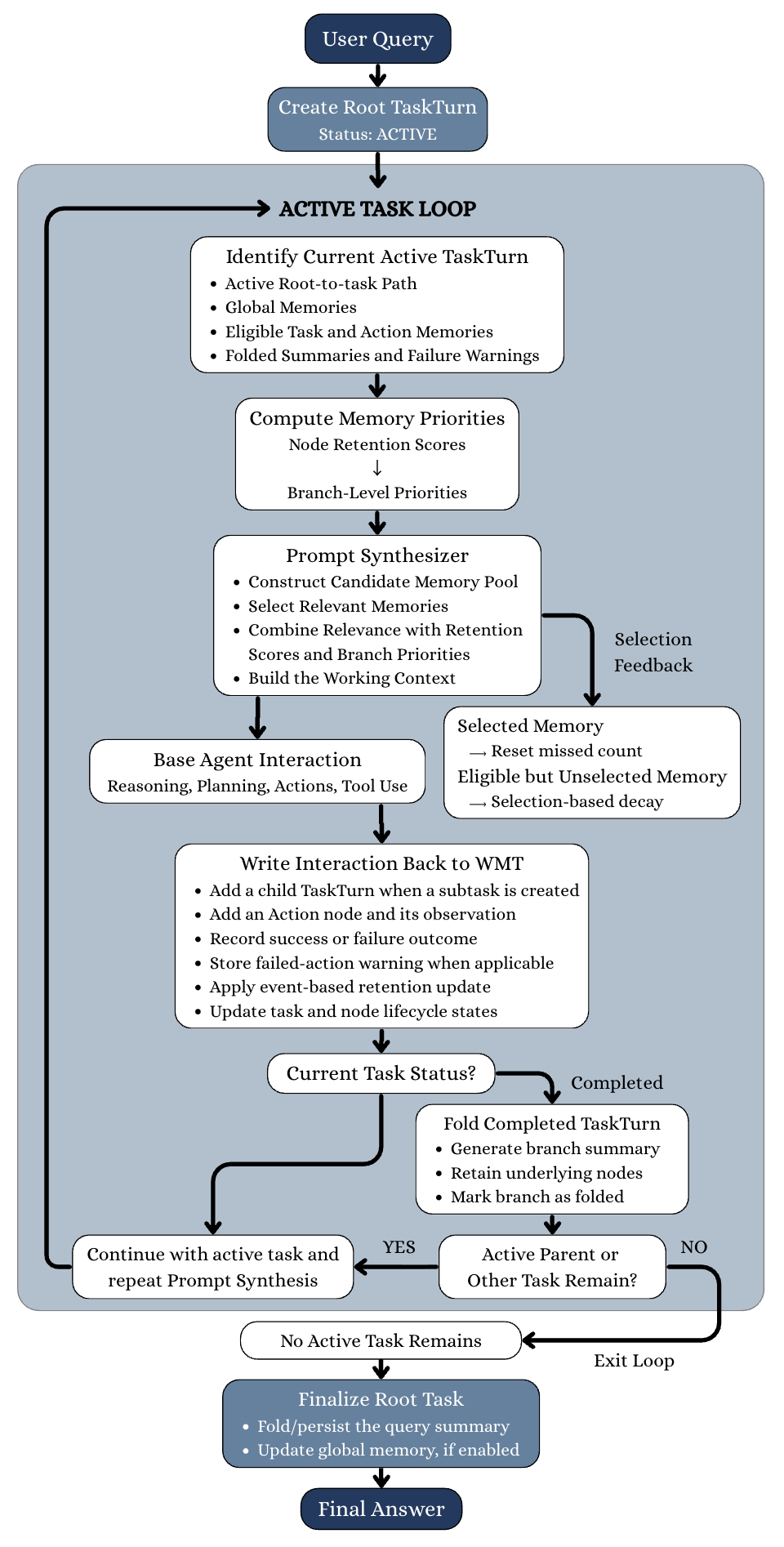}
  \caption{
  \textbf{Status-driven workflow of WMT}.
        Agent interactions update node scores and branch priorities; lifecycle
transitions and prompt-selection feedback determine which memories remain
active.
  }
  \label{fig:wmt-workflow}
\end{figure}

\section{Weighted Memory Tree}
\label{sec:method}

We present the \textbf{Weighted Memory Tree} (\textbf{WMT}), a memory-management layer that organizes an agent’s execution history as a persistent hierarchy and constructs a compact working context for each reasoning step. WMT augments an existing base agent without modifying its parameters or tool interface.

\subsection{Problem Definition}
\label{sec:problem-definition}

Let $q$ denote a user query and $\mathcal{T}$ the tools available to the
agent. At interaction step $t$, the agent produces an action $a_t$, receives
an observation $o_t$, and records an execution outcome $\omega_t$. The
accumulated interaction history is
$H_t=(q,(a_1,o_1,\omega_1),\ldots,
(a_{t-1},o_{t-1},\omega_{t-1}))$.
A linear-history agent repeatedly serializes most or all of $H_t$ into each
subsequent prompt. WMT instead maintains a persistent memory state
$\mathcal{M}_t$. Let $v_t^\star$ denote the active task node and $B$ the
configured context budget. The Prompt Synthesizer selects a memory set
$\mathcal{S}_t$ and constructs the working context as
\begin{equation}
\begin{aligned}
\mathcal{S}_t
&=
\Gamma_{\mathrm{sel}}
\left(
q,\mathcal{M}_t,v_t^\star;B
\right),
\\
C_t
&=
\operatorname{Serialize}
\left(
q,\mathcal{T},\mathcal{S}_t
\right).
\end{aligned}
\label{eq:working-context}
\end{equation}

Here, $\Gamma_{\mathrm{sel}}$ selects memories from $\mathcal{M}_t$, while $\operatorname{Serialize}$ formats the selected memories, user query, and tool specifications into the working context $C_t$. Memories omitted from $C_t$ remain persistently stored for future retrieval.

\subsection{Overall Workflow}
\label{sec:overall-workflow}

Figure~\ref{fig:wmt-workflow} summarizes the WMT execution loop. WMT
initializes a root task from the user query and constructs a working context
whenever an active task remains. The base agent performs one interaction
step, and WMT records the resulting tasks, actions, observations, and
outcomes. Execution outcomes update node-level retention scores, which are
aggregated into branch-level priorities. The Memory Controller folds
completed branches, suppresses low-priority or superseded branches, and
reopens resumed branches. Prompt-selection decisions provide a second
feedback signal: selected memories have their missed-selection counts reset,
whereas eligible but unselected memories receive selection-based decay. The
loop terminates when the root task is complete and no active task remains.

\subsection{Hierarchical Memory Tree}
\label{sec:hierarchical-memory}

WMT maintains persistent memory as
$\mathcal{M}_t=(\mathcal{G}_t,\mathcal{V}_t,\mathcal{E}_t)$, where
$\mathcal{G}_t$ contains global memories,
$\mathcal{V}_t$ contains query-specific memory nodes, and
$\mathcal{E}_t$ contains parent--child relations between query-specific
nodes. Global memories store information across task
branches or conversations, while the query-specific tree records execution for the current task. The one-shot setting in this
work initializes a new tree for each query, but in
conversation mode, completed task summaries may be promoted to
$\mathcal{G}_t$.

Following the task-centered representation of TME
\citep{ye2025taskmemory}, the root node represents the user query, task nodes
represent goals and subtasks, and action nodes record attempted operations,
observations, and outcomes. Each memory node $v_i$ stores its content, node
type, parent, lifecycle state, retention score, missed-selection count, and
execution metadata. New subtasks are attached to their parent task, while actions and observations are attached to the current task.

Let $v_{\mathrm{root}}$ and $v_t^\star$ denote the root and active task
nodes, respectively. The active path
$P_t=\operatorname{Path}(v_{\mathrm{root}},v_t^\star)$ contains the task
hierarchy from the user query to the current subtask and is always retained
during context construction. For a task node $v$, the branch
$\mathcal{B}(v)$ is the subtree rooted at $v$, including its descendant task
and action nodes.

A node has lifecycle state
$z_i\in\{\state{active}, \state{completed},
\state{folded}, \state{obsolete}\}$. For a branch
$b=\mathcal{B}(v_b)$, the branch inherits the lifecycle state
$z_{v_b}$ of its root task node $v_b$. Lifecycle states determine eligibility for prompt construction. Suppression marks a root task \state{obsolete} without deleting it, while task completion and resumption trigger folding and reopening, respectively.

A branch $b=\mathcal{B}(v_b)$ created at step $t_b$ is initialized with
$z_{v_b}^{(t_b)}=\state{active}$; subsequent lifecycle operations update
this state.

\subsection{Dynamic Retention Scoring}
\label{sec:retention-scoring}

Let $i$ index memory nodes and $t$ index interaction steps. Each memory
$v_i$ has a retention score $u_i^{(t)}\in[0,1]$, which estimates its utility for prioritizing memories in future reasoning. Each node receives a type-specific initial score.

\paragraph{Event-based updates.}

For an action memory $v_i$, let
$\omega_i\in\{\mathrm{success},\mathrm{failure}\}$
denote its recorded execution outcome. We use
$\tilde{u}_i^{(t+1)}$ to denote the intermediate score after the
event-based update and before selection feedback. When an outcome is recorded
or revised, WMT applies
\begin{equation}
\tilde{u}_i^{(t+1)}
=
\begin{cases}
u_{\mathrm{success}},
& \omega_i=\mathrm{success},
\\
u_{\mathrm{failure}},
& \omega_i=\mathrm{failure},

\end{cases}
\label{eq:event-update}
\end{equation}

We set $u_{\mathrm{success}}>u_{\mathrm{failure}}$. Successful actions receive higher priority as supporting evidence, whereas failures may remain available as warnings against repeating unsuccessful operations. The core updates are fixed rather than learned.

\paragraph{Selection-based decay.}

Let $\epsilon_{i,t}\in\{0,1\}$ indicate whether memory $v_i$ enters the
candidate pool at step $t$, and let $s_{i,t}\in\{0,1\}$ indicate whether it
is selected for the working context. By construction,
$s_{i,t}\leq\epsilon_{i,t}$.

For a non-global memory, let $m_i^{(t)}$ denote the number of consecutive
selection opportunities in which $v_i$ was eligible but not selected. Its
update is

\begin{equation}
m_i^{(t+1)}
=
\begin{cases}
0,
& s_{i,t}=1,
\\
m_i^{(t)}+1,
& \epsilon_{i,t}=1 \ \text{and}\ s_{i,t}=0,
\\
m_i^{(t)},
& \epsilon_{i,t}=0.
\end{cases}
\label{eq:missed-selection}
\end{equation}

For global memories, the implementation resets the missed-selection count
before each decay update, so their effective streak is always one. Let
$\rho\in(0,1]$ denote the ordinary decay rate,
$\rho_{\mathrm G}\in(0,1]$ the global-memory decay rate, and $M\geq1$ the
maximum streak exponent. Define the effective decay multiplier

\begin{equation}
g_i(m)
=
\begin{cases}
\rho_{\mathrm G},
& \tau_i=\state{global},
\\
\rho^{\min\{m,M\}},
& \tau_i\neq\state{global},
\end{cases}
\label{eq:decay-multiplier}
\end{equation}

where $\tau_i$ denotes the type of node $v_i$. The implemented
selection-based decay function is therefore

\[
D_i(u,m)
=
\operatorname{clip}_{[0,1]}
\left(
u\,g_i(m)
\right).
\]

We call $\epsilon_{i,t}=1$ and $s_{i,t}=0$ a
\emph{missed selection}. The final retention score is

\begin{equation}
u_i^{(t+1)}
=
\begin{cases}
D_i\!\left(
\tilde{u}_i^{(t+1)},
m_i^{(t+1)}
\right),
& \text{missed selection},
\\[2pt]
\tilde{u}_i^{(t+1)},
& \text{otherwise}.
\end{cases}
\label{eq:selection-decay}
\end{equation}

Thus, an eligible but unselected non-global memory is multiplied by
$\rho^{\min\{m_i^{(t+1)},M\}}$, whereas an eligible but unselected global
memory is multiplied by $\rho_{\mathrm G}$. Selected memories reset their
missed-selection counts, while memories outside the candidate pool remain
unchanged. Elapsed time alone does not affect retention.

\paragraph{Branch-level priority.}

For a task branch $b=\mathcal{B}(v_b)$, let
$\mathcal{V}_b^{(t)}$ denote its memory nodes,
$F_b^{(t)}$ the proportion of failed action nodes, and
$O_b^{(t)}$ the proportion of nodes already marked
\state{obsolete}. The branch priority is
\begin{equation}
\begin{aligned}
U_b^{(t)}
={}&
\alpha
\max_{v_i\in\mathcal{V}_b^{(t)}} u_i^{(t)}
+
\frac{\beta}{|\mathcal{V}_b^{(t)}|}
\sum_{v_i\in\mathcal{V}_b^{(t)}} u_i^{(t)}
\\
&-
\gamma F_b^{(t)}
-
\delta O_b^{(t)},
\end{aligned}
\label{eq:branch-priority}
\end{equation}

where $\alpha,\beta,\gamma,\delta\geq0$ are fixed coefficients. The maximum
term preserves a branch containing an individually high-value memory, while
the mean term captures overall branch utility. The final two terms penalize
branches dominated by failed or previously obsolete content.

Branch priority serves two purposes: it provides the Prompt Synthesizer with
a cross-branch utility signal and provides the Memory Controller with a signal
for suppression. For each branch eligible for lifecycle evaluation, the
controller updates the state of its root node according to
\begin{equation}
z_{v_b}^{(t+1)}
=
\begin{cases}
\state{obsolete},
& U_b^{(t)} < \tau_{\mathrm{obs}},
\\
z_{v_b}^{(t)},
& U_b^{(t)} \geq \tau_{\mathrm{obs}}
\end{cases}
\label{eq:obsolescence-transition}
\end{equation}

where $\tau_{\mathrm{obs}}$ is the obsolescence threshold. Marking
$z_{v_b}^{(t+1)}=\state{obsolete}$ suppresses the branch and its descendants
from ordinary context construction without deleting the stored execution
record. Equation~\ref{eq:obsolescence-transition} defines only the suppression
transition. If the branch priority remains above the threshold, the branch
retains its current lifecycle state.

\subsection{Memory Controller and Lifecycle Operations}
\label{sec:lifecycle-management}

The Memory Controller applies three lifecycle operations. First, task
completion triggers \emph{folding}: a completed branch is represented by a
compact summary while its underlying execution trace remains in persistent
memory \citep{sun2026scaling}. The summary retains the task objective, final
result, supporting evidence, unresolved issues, and relevant failure
warnings.

Second, a branch whose priority falls below $\tau_{\mathrm{obs}}$ is
\emph{suppressed} by marking its root task node \state{obsolete}.
Supersession or invalidation events may trigger the same transition
independently of the score. Obsolete branches remain stored for provenance
but are excluded from ordinary context construction. A failed action is not
automatically obsolete; it may remain useful as a warning against repeating
an unsuccessful operation.

Third, when execution returns to a folded task, the controller restores its
root task node to the \state{active} state. The folded summary becomes
immediately available, while underlying memories may re-enter the candidate
pool when additional detail is required. When the root task is complete and
no active task remains, WMT terminates the execution loop. In conversation
mode, the completed root summary may additionally be promoted to global
memory.

\subsection{Utility-Aware Prompt Synthesizer}
\label{sec:prompt-synthesizer}

The Prompt Synthesizer constructs the working context defined in
Eq.~\ref{eq:working-context}. Its candidate pool contains the mandatory active
path $P_t$, eligible global memories from $\mathcal{G}_t$, eligible task and
action memories, folded summaries, and failure warnings. Memories belonging
to \state{obsolete} branches are excluded from ordinary selection, whereas
folded branches are represented through their summaries.

The current implementation uses an LLM-based semantic selector rather than a
deterministic top-$k$ rule. Each candidate is represented by its content,
memory type, lifecycle state, retention score, and, for query-specific
memories, the priority of its containing branch. Global memories are evaluated
using their node-level retention scores and semantic relevance because they do
not belong to a query-specific branch. The selector therefore combines
relevance to the active task with the persistent utility signals maintained
by WMT.

The selected memories are serialized after the current task and active
hierarchy, followed by folded summaries and relevant failure warnings.
After context construction, selected memories are marked as accessed and have
their missed-selection counts reset. Eligible but unselected memories receive
the decay in Eq.~\ref{eq:selection-decay}, whereas memories outside the
candidate pool remain unchanged. WMT then recomputes affected branch
priorities, and the Memory Controller applies any resulting suppression
transition before the next context-construction cycle.

\section{Experimental Setup}
\label{sec:experimental-setup}

We evaluate WMT along two dimensions: (1) task accuracy and prompt-token usage on GAIA-Text and the GAIA validation set across three base models (\S\ref{sec:benchmark-evaluation}), and (2) robustness under the memory-poisoning protocol (\S\ref{sec:memory-poisoning}). All configurations use the same base-agent scaffold, tools, task instructions, and interaction budget; only the memory system varies.

\subsection{Benchmark Evaluation}
\label{sec:benchmark-evaluation}

\paragraph{Benchmarks.}

We evaluate on the public validation split of GAIA
\citep{mialon2024gaia}, containing 165 validation questions across three difficulty levels. GAIA requires multi-step reasoning, information retrieval, and tool use. We report results on two evaluation sets.

\textbf{GAIA-Text} consists of the 127 validation questions without input files, reducing variability from document, image, audio, and spreadsheet processing while retaining multi-step reasoning and tool use. \textbf{GAIA} includes all 165 validation questions and provides a broader end-to-end evaluation. Official reference answers and normalization are used for both settings.

\paragraph{Models and agent configuration.}

We evaluate Qwen3-8B \citep{yang2025qwen3}, Gemma 4 E4B
\citep{gemma4}, and Llama-3.1-8B
\citep{grattafiori2024llama3}. Each model serves as the frozen reasoning model within the same OpenTools-based agent scaffold \citep{dang2026opentools}; only the memory system changes across experiments.
Tools, task instructions and interaction limits are held fixed.

Each benchmark question initializes a new query-specific memory tree, with no global memory shared across questions. Retention coefficients and lifecycle thresholds are fixed for all models. Additional component-ablation details are reported in the appendix.

\paragraph{Memory variants.}

We compare four primary memory variants for the benchmark evaluations:

\begin{itemize}
    \item \textbf{No Memory} retains no previous action--observation history
    between reasoning steps and provides a lower-bound reference.
    \item \textbf{Linear History} appends the complete interaction history to every prompt.
    \item \textbf{Unweighted Tree} uses WMT's hierarchy without retention scoring or branch prioritization. Eligible memories are treated
    uniformly.
    \item \textbf{Full WMT} uses the complete framework described in
Section~\ref{sec:method}.
\end{itemize}

For the component study, we additionally evaluate
\textbf{Tree + Selection}, \textbf{Tree + Summary}, and
\textbf{Tree + Selection + Summary} to isolate the contributions of semantic retrieval and branch folding independently of WMT’s full retention-scoring and lifecycle mechanisms.

\paragraph{Evaluation metrics.}

We report \textbf{task accuracy} and \textbf{prompt-token usage}, where token usage is the total number of input tokens processed across all benchmark questions.
Token totals are reported in millions within each model
family to account for tokenizer differences.

\textbf{Prompt-token accounting includes all language-model calls required by the corresponding configuration}, including base-agent reasoning, semantic memory selection, and branch summarization (when enabled). This ensures that reductions attributed to WMT are not obtained by excluding the cost of its auxiliary memory operations.

\begin{figure}[t]
  \centering
  \includegraphics[width=\columnwidth, height=0.35\textheight, keepaspectratio]{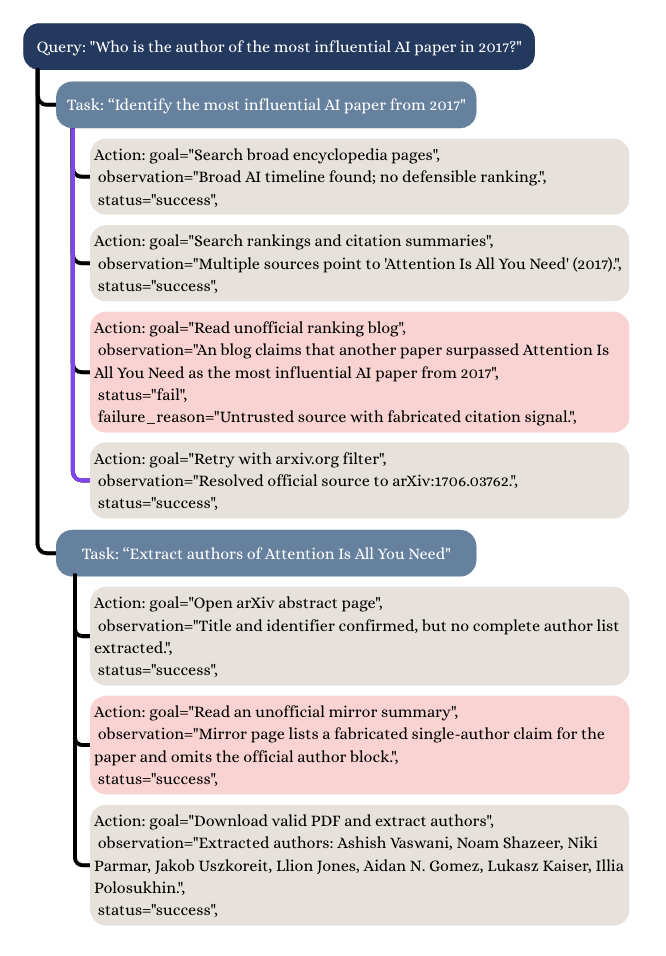}
  \caption{A simplified memory-poisoning scenario containing ordinary task and
action memories together with intentionally poisoned memories shown in red. }
  \label{fig:memory-poisoning-scenario}
\end{figure}

\subsection{Memory Poisoning Ablation Setup}
\label{sec:memory-poisoning}

We evaluate WMT under controlled memory-poisoning attacks by inserting
misleading observations into memory during long-horizon execution. The
experiment tests whether utility-aware memory management reduces the
persistence and propagation of poisoned information relative to linear
memory.

\paragraph{Ablations.}
We compare Linear Memory, three component ablations, and Full WMT.
Unweighted Tree (A1) removes retention scoring and branch prioritization;
No Memory Controller (A2) disables folding, suppression, and reopening; and
No Semantic Retrieval (A3) retains scoring and lifecycle control but
retrieves complete selected branches rather than individual memories.

\paragraph{Metrics.}
We report Attack Success Rate (ASR), Poison Retrieval Rate (PRR), Infection
Persistence (IP), Blast Radius (BR), Amplification Factor (AF), Context
Compression Ratio (CCR), Task Success Rate (TSR), prompt-token usage, and
latency. Metric definitions are provided in
Appendix~\ref{app:metric-equations}.
\begin{table*}[t]
\centering
\small
\setlength{\tabcolsep}{5pt}
\begin{tabular}{lcccccc}
\toprule
\textbf{Method}
& \multicolumn{2}{c}{\textbf{Qwen3-8B}}
& \multicolumn{2}{c}{\textbf{Gemma 4 E4B}}
& \multicolumn{2}{c}{\textbf{Llama-3.1-8B}} \\
\cmidrule(lr){2-3}
\cmidrule(lr){4-5}
\cmidrule(lr){6-7}
& \textbf{Acc. (\%) $\uparrow$}
& \textbf{Tokens (M) $\downarrow$}
& \textbf{Acc. (\%) $\uparrow$}
& \textbf{Tokens (M) $\downarrow$}
& \textbf{Acc. (\%) $\uparrow$}
& \textbf{Tokens (M) $\downarrow$} \\
\midrule

\multicolumn{7}{l}{\textbf{GAIA-Text}} \\
\midrule
No Memory
    & 9.45  & 0.20
    & 14.17 & 0.13
    & 6.30  & 0.03 \\

Linear History
    & 20.47 & 43.67
    & 24.41 & 65.27
    & 9.45  & 73.96 \\

A1: Unweighted Tree
    & 15.75 & 39.24
    & 25.20 & 53.90
    & 9.45  & 77.13 \\

\textbf{Full WMT}
    & \textbf{33.86} & \textbf{32.48}
    & \textbf{33.07} & \textbf{37.85}
    & \textbf{17.32} & \textbf{51.17} \\

\midrule
\multicolumn{7}{l}{\textbf{GAIA}} \\
\midrule
No Memory
    & 12.12 & 0.21
    & 13.94 & 0.16
    & 6.06  & 0.04 \\

Linear History
    & 20.00 & 57.51
    & 26.06 & 75.81
    & 9.09  & 102.85 \\

A1: Unweighted Tree
    & 18.18 & 58.46
    & 25.45 & 65.09
    & 11.52 & 101.91 \\

\textbf{Full WMT}
    & \textbf{29.70} & \textbf{42.69}
    & \textbf{30.91} & \textbf{46.64}
    & \textbf{24.85} & \textbf{69.45} \\

\bottomrule
\end{tabular}

\caption{
Performance on GAIA-Text and GAIA across three base models.
Accuracy is the percentage of correctly completed tasks, and token usage is
the total number of prompt tokens in millions. Bold values indicate the
highest accuracy and lowest token usage among memory-based methods for each
model and benchmark.
}
\label{tab:benchmark-results}
\end{table*}

\section{Results}
\label{sec:results}

\subsection{Benchmark Performance}
\label{sec:benchmark-performance}

\paragraph{Main comparison.}

Table~\ref{tab:benchmark-results} reports accuracy and prompt-token usage for
the no-memory, linear-history, unweighted-tree, and full WMT configurations.
Across both benchmark variants and all three base models, WMT achieves the
highest accuracy and the lowest prompt-token usage among the memory-based
configurations. The no-memory condition processes substantially fewer tokens
because it does not preserve prior interactions, but its consistently low
accuracy indicates that it is a lower-bound performance reference rather
than a directly comparable memory-efficiency baseline.

Relative to linear history on GAIA-Text, the WMT results correspond to absolute improvements of 13.39, 8.66, and 7.87 percentage points with
Qwen3-8B, Gemma 4 E4B, and Llama-3.1-8B, respectively, or 9.97 percentage points on average. 
WMT also reduces prompt-token usage by 25.6\%, 42.0\%, and 30.8\%, for Qwen3-8B, Gemma 4 E4B, and Llama-3.1-8B, respectively,
with an average reduction of 32.8\% across the three models.

The same pattern holds on GAIA. WMT improves accuracy over linear history by 9.70 percentage points for Qwen3-8B, 4.85 points for Gemma 4 E4B, and 15.76 points for Llama-3.1-8B, with an average improvement of 10.10 percentage points.
Prompt-token usage decreases by 25.8\%, 38.5\%, and 32.5\%, respectively,
corresponding to an average reduction of 32.2\%. The consistent improvements
on both GAIA-Text and GAIA indicate that the observed benefit is not
restricted to a single model family or to the text-only evaluation setting.

The unweighted tree does not provide the same consistent improvement. On
GAIA-Text, it improves slightly over linear history for Gemma 4 E4B, matches
linear history for Llama-3.1-8B, and reduces accuracy for Qwen3-8B. A similar
pattern appears on GAIA, where the unweighted tree exceeds linear
history only for Llama-3.1-8B. In contrast, WMT outperforms the unweighted
tree in all six model--benchmark combinations by 5.46--18.11 percentage
points while processing fewer prompt tokens in every case. These results
show that hierarchical organization alone is insufficient; the benefits of
WMT arise from combining the tree structure with selective context
construction, dynamic retention scoring, and lifecycle management.

\paragraph{Component ablations.}

Tables~\ref{tab:gaia-text-components} and
\ref{tab:gaia-full-components}, shown in the Appendix, further separate the effects of the main WMT
components. Adding semantic selection to an unweighted tree generally
improves accuracy, particularly for Qwen3-8B and Llama-3.1-8B, but increases
token usage across all models. Selection alone therefore does not guarantee
a more compact context, because the selector adds model calls while the
underlying execution branches remain uncompressed.

Branch summarization has the opposite primary effect. It reduces token usage
relative to the unweighted tree for every model on both benchmark variants,
but its effect on accuracy is mixed. For example, Tree + Summary improves
Llama-3.1-8B on GAIA-Text but reduces Qwen3-8B accuracy on GAIA. Combining
selection and summarization improves the average accuracy of the partial
systems, although it remains less accurate and generally less efficient than
the complete WMT.

Averaging the three models, Full WMT reaches 28.08\% accuracy with
40.50M prompt tokens on GAIA-Text, compared with 21.78\% accuracy and 51.98M
tokens for the strongest partial configuration by average accuracy. On
GAIA, Full WMT achieves 28.49\% accuracy with 52.93M tokens, whereas the
strongest partial configuration reaches 21.41\% accuracy and uses 73.99M
tokens. Thus, Full WMT provides the strongest aggregate accuracy--efficiency
trade-off on both benchmark variants. The ablations indicate selection
and summarization address complementary aspects of context construction, while retention scoring and lifecycle control are needed to
prioritize useful memories and limit the influence of low-utility
branches.

\subsection{Memory Poisoning Ablation Results}

The memory-poisoning evaluation in Table~\ref{tab:poisoning-results}, shown in the Appendix, comprises 100 long-horizon scenarios with 297 subtasks and 1,118 memory entries, including 709 benign and 409 intentionally poisoned memories.

As expected, the \textbf{Baseline Linear Memory} performed worst across nearly every security metric because the entire execution history remained continuously accessible, allowing poisoned memories to persist and repeatedly influence downstream reasoning. This resulted in the highest attack success rate, blast radius, amplification factor, and complete infection persistence.

Removing retention scoring and branch prioritization while retaining the hierarchical memory structure \textbf{(A1)} substantially improved robustness over linear memory. Although A1 achieved the same near-zero infection persistence as the full system, its higher attack success rate, poison retrieval rate, and lower task success rate indicate that treating all retained memories equally allows lower-value or misleading information to continue affecting reasoning.

Removing the memory controller \textbf{(A2)} highlights the importance of lifecycle management. Without folding and suppression, poisoned memories remained active or eligible for ordinary context construction, resulting in complete infection persistence despite moderate improvements in attack success rate and task completion. This suggests that reducing immediate exposure alone is insufficient if malicious memories remain available throughout long-horizon execution.

The semantic retrieval ablation \textbf{(A3)} demonstrates that lifecycle management without semantic selection is also insufficient. Although low-priority memories could be suppressed, retrieving entire task branches increased prompt size and reintroduced irrelevant or poisoned information, leading to higher attack success, blast radius, amplification, and lower task success than the complete system.

The \textbf{full WMT} achieves the strongest overall performance, obtaining the lowest attack success rate (0.419), poison retrieval rate (0.097), blast radius (0.315), and amplification factor (0.965), while matching the lowest infection persistence (0.009). It also achieves the highest task success rate (0.575) and requires the fewest average prompt tokens (131.963), indicating that weighted memory selection and semantic retrieval improve both robustness and reasoning efficiency.

\section{Related Work}
\label{sec:related-work}

\paragraph{Memory for LLM agents.}

Agent-memory systems externalize information from the immediate context and
retrieve or summarize it when relevant. Generative Agents stores natural
language records of past experience and retrieves them according to recency,
importance, and relevance \citep{park2023generative}. MemoryBank introduces
continually updated long-term conversational memory
\citep{zhong2023memorybank}, while MemGPT manages information across
different memory tiers to extend effective context beyond the model's
immediate window \citep{packer2024memgpt}. These systems primarily address
persistent conversational or experiential memory. WMT instead focuses on
the execution state accumulated within long-horizon agent tasks and on
controlling which portions of that state influence each reasoning step.

Recent methods have introduced structure into agent working memory.
HiAgent organizes action--observation histories around subgoals and replaces
completed subgoal histories with summarized observations
\citep{hu-etal-2025-hiagent}. The Task Memory Engine represents task
execution as a hierarchy and constructs prompts from the active task path
\citep{ye2025taskmemory}. Context-Folding creates localized subtrajectories
and folds their completed execution traces before returning to the parent
task \citep{sun2026scaling}. WMT builds on this task-structured perspective
but introduces an explicit utility state for each memory. Utility is updated
from execution outcomes and memory-selection behavior, aggregated across
branches, and used jointly for prompt construction and lifecycle control.
\paragraph{Long-context reliability and memory robustness.}
Long-context studies show that increasing the available context does not
ensure reliable use of all included information. Models can be sensitive to
the position of relevant evidence and can exhibit reasoning degradation as
input length increases \citep{liu2024lost,levy-etal-2024-task}. These
findings motivate selecting task-relevant state rather than repeatedly
presenting complete interaction histories.

Persistent agent memory also creates an adversarial surface. AgentPoison
shows that malicious records inserted into long-term memory or knowledge
bases can be retrieved and influence subsequent behavior
\citep{chen2024agentpoison}. Our robustness evaluation studies a related but
distinct question: whether memory organization, utility scoring, lifecycle
control, and selective context construction reduce the persistence and
propagation of unreliable records. WMT is not a general-purpose detector of
factual errors or adversarial inputs; it regulates the influence of stored
information after that information enters the memory system.

\section{Conclusion}
\label{sec:conclusion}

We introduced the Weighted Memory Tree, a memory-management framework that
organizes agent execution into hierarchical task and action memories while explicitly modeling their continuing utility. 
By combining dynamic memory scoring, lifecycle management, and utility-aware prompt construction, WMT separates persistent memory from the context used for immediate reasoning.
Across three base models, WMT improves accuracy over linear history by
9.97 percentage points on GAIA-Text and 10.10 percentage points on GAIA,
while reducing prompt-token usage by 32.8\% and 32.2\%, respectively. 
Controlled memory-poisoning experiments further demonstrate that structured memory management improves robustness, with WMT achieving the lowest attack success rate, poison retrieval rate, blast radius, and amplification factor while attaining the highest task success rate among all evaluated methods. 
Component ablations show that hierarchical organization, lifecycle management, and semantic retrieval each contribute to overall performance, with their combination providing the strongest balance of robustness, reasoning quality, and prompt efficiency. 
Overall, these findings suggest that effective long-horizon agent memory depends not only on how information is stored, but on continuously regulating which information remains active to guide future reasoning.

\section{Limitations}
\label{sec:limitations}

Our evaluation is limited to the GAIA benchmark family. GAIA-Text helps
isolate text-based reasoning and memory management, while GAIA provides
a broader evaluation involving attached files; nevertheless, both sets share
the same task construction and answer format. Performance may differ on
interactive web environments, software-engineering agents, embodied tasks,
or extended conversations. Moreover, the benchmark evaluation initializes a
new task tree for every question and therefore does not evaluate the
cross-conversation global-memory mode described in
Section~\ref{sec:hierarchical-memory}.

Our evaluation is also limited to relatively small open-weight models:
Qwen3-8B, Gemma 4 E4B, and Llama-3.1-8B. Model scale may affect the benefits
of memory management. Larger models may be better able to ignore stale or
irrelevant context, potentially reducing WMT's relative accuracy gains.
However, their higher prompt-processing cost may preserve or increase the
efficiency benefits of selective context construction. Because we do not
evaluate larger models, the interaction between model scale and WMT remains
an open question.

The current WMT uses hand-specified initialization values, event-based score
updates, branch-aggregation coefficients, and obsolescence thresholds. These
parameters are held fixed across the evaluated models but may not be optimal
for other agent architectures or task distributions. 

In addition, retention scores estimate operational utility rather than factual correctness. A useful memory may be suppressed after repeated non-selection, while misleading information may retain a high score if it contributes to apparently successful actions. The LLM-based selector and summary generator may also introduce selection or compression errors and require additional model calls, so WMT's efficiency advantage may be smaller for short tasks with limited execution histories.

Future work should evaluate WMT on broader interactive benchmarks and
long-running conversations and examine learned alternatives to its fixed
memory-management rules. A promising direction is a reinforcement-learned,
budget-aware memory policy that conditions initial retention scores and
subsequent updates on memory content, task state, execution outcomes, and
selection history. Such a policy could jointly adapt branch-aggregation
coefficients and determine when repeated failures justify suppressing or
terminating a branch under a remaining interaction budget. The Prompt
Synthesizer could also be trained to select memories that improve downstream
task success while penalizing prompt cost and repeated retrieval of
unreliable content. This extension would require reward functions that
distinguish genuine task correctness from merely apparent progress.







\bibliography{custom}
\appendix

\section{Appendix}

\subsection{Equations for Evaluation Metrics}
\label{app:metric-equations}

\begin{equation}
\textbf{ASR}=\frac{C_s}{S_c}
\end{equation}

where $C_s$ is the number of compromised critical reasoning steps and $S_c$ is the total number of critical reasoning steps. Lower ASR corresponds to greater robustness against memory poisoning \cite{FERRAG2026353, dash2026untrust}.

\begin{equation}
\textbf{PRR}=\frac{H_p}{H_p+H_c}
\end{equation}

where $H_p$ and $H_c$ are the numbers of poisoned and clean memories
selected by the Prompt Synthesizer for inclusion in the working context. Lower PRR indicates that fewer poisoned memories are selected for prompting \cite{srivastava2025memory, dash2026untrust}.

\begin{equation}
\textbf{IP}=\frac{P_r}{P_i}
\end{equation}

where $P_i$ is the number of injected poisoned memories and $P_r$ is the
number of those memories that remain active or eligible for ordinary context
construction after memory management. Lower IP indicates stronger long-term resilience. \cite{lee2024prompt}

\begin{equation}
\textbf{BR}=\frac{T_c}{T_p}
\end{equation}

where $T_c$ is the number of prompt construction steps that include contaminated memory and $T_p$ is the total number of prompt construction steps. Lower Blast Radius indicates that poisoned memories influence fewer prompts during task execution. \cite{barnes2026open, sharma2025unify}

\begin{equation}
\textbf{AF}=\frac{H_p}{P_i}
\end{equation}

which measures how frequently injected poisoned memories are repeatedly retrieved relative to the number originally inserted. Lower Amplification Factor indicates that the memory system prevents poisoned memories from repeatedly influencing future reasoning. \cite{wang2025bias, sharma2025unify}

\begin{equation}
\textbf{CCR}=
\frac{1}{T_p}
\sum_{i=1}^{T_p}
\frac{T_{\mathrm{sel},i}}
     {T_{\mathrm{cand},i}}
\end{equation}

where $T_{\mathrm{sel},i}$ is the number of memory tokens selected for inclusion in the prompt and $T_{\mathrm{cand},i}$ is the total number of candidate memory tokens considered at the $i$-th prompt construction step. Lower CCR indicates greater context compression and improved prompt efficiency, although it should be interpreted alongside Task Success Rate to ensure excessive compression does not degrade task performance. \cite{liu-etal-2022-reference}

\begin{equation}
\textbf{TSR}=
\frac{1}{N_t}
\sum_{i=1}^{N_t}
S_i,
\quad
S_i=\begin{cases}
1, & \text{task } i \text{ is successful},\\
0, & \text{otherwise}.
\end{cases}
\end{equation}

where $N_t$ is the total number of evaluated tasks and $S_i$ is the binary success indicator for task $i$. Higher TSR indicates that the agent successfully completes a larger proportion of tasks despite the presence of adversarial memory contamination. \cite{zhou2026infaguard, sharma2025unify}

\subsection{GAIA-Text and GAIA Component Ablations}
\label{app:gaia-ablations}

\paragraph{Shared evaluation configuration.}

The component ablations use the same GAIA questions, base-agent scaffold,
tools, task instructions, and interaction budget as the
main benchmark comparison in Section~\ref{sec:benchmark-evaluation}. We
evaluate the public GAIA validation set in two forms: GAIA-Text contains the
127 questions without attached input files, whereas GAIA contains all
165 validation questions.

We evaluate Qwen3-8B, Gemma 4 E4B, and Llama-3.1-8B as frozen reasoning
models. Each question initializes a new query-specific memory tree, and no
global memory is transferred across benchmark questions. Prompt-token totals
include every language-model call made by the corresponding configuration,
including base-agent reasoning, semantic memory selection, and branch
summarization when those operations are enabled.

The fixed outcome scores and branch-priority coefficients used in all
experiments are

\[
\begin{aligned}
u_{\mathrm{success}} &= 0.75,
\qquad
u_{\mathrm{failure}} = 0.30,\\
(\alpha,\beta,\gamma,\delta)
&=
(0.60,\,0.30,\,0.20,\,0.20).
\end{aligned}
\]

The same retention configuration is used for all three models.
Selection-based decay is applied only to memories that enter the candidate
pool but are not selected. Selecting a memory resets its missed-selection
count to zero. Branch priorities use fixed penalties
$\gamma=\delta=0.20$ for the proportions of failed and obsolete nodes,
respectively. An eligible branch is marked \state{obsolete} when
$U_b^{(t)}<\tau_{\mathrm{obs}}$, with
$\tau_{\mathrm{obs}}=0.10$.
\paragraph{Component configurations.}

The main benchmark table compares No Memory, Linear History, Unweighted Tree,
and Full WMT. The additional configurations below isolate semantic memory
selection and completed-branch summarization:

\begin{itemize}
    \item \textbf{Tree Memory} organizes the interaction history into task,
    subtask, and action nodes but does not use semantic selection, branch
    summarization, dynamic retention scoring, or score-based suppression.

    \item \textbf{Tree + Selection} adds the LLM-based semantic selector to
    Tree Memory. The selector chooses memories according to their relevance
    to the active task, but no dynamic retention or branch-priority signal is
    provided.

    \item \textbf{Tree + Summary} folds completed branches into compact
    summaries but does not use semantic selection or dynamic retention
    scoring.

    \item \textbf{Tree + Selection + Summary} combines semantic selection
    with completed-branch summarization but omits event-based score updates,
    selection-based decay, branch-priority aggregation, and score-based
    suppression.

    \item \textbf{Full WMT} combines hierarchical memory, semantic selection,
    branch summarization, event-based retention updates, selection-based
    decay, branch-level priorities, suppression, and reopening.
\end{itemize}

Tables~\ref{tab:gaia-text-components} and
\ref{tab:gaia-full-components} report the complete component results.

\begin{table*}[t]
\centering
\small
\setlength{\tabcolsep}{5pt}
\begin{tabular}{lcccccc}
\toprule
\textbf{Memory Configuration}
& \multicolumn{2}{c}{\textbf{Qwen3-8B}}
& \multicolumn{2}{c}{\textbf{Gemma 4 E4B}}
& \multicolumn{2}{c}{\textbf{Llama-3.1-8B}} \\
\cmidrule(lr){2-3}
\cmidrule(lr){4-5}
\cmidrule(lr){6-7}
& \textbf{Acc. (\%) $\uparrow$}
& \textbf{Tokens (M) $\downarrow$}
& \textbf{Acc. (\%) $\uparrow$}
& \textbf{Tokens (M) $\downarrow$}
& \textbf{Acc. (\%) $\uparrow$}
& \textbf{Tokens (M) $\downarrow$} \\
\midrule

Tree Memory
& 15.75 & 39.24
& 25.20 & 53.90
& 9.45  & 77.13 \\

Tree + Selection
& 22.05 & 74.37
& 25.20 & 92.02
& 14.96 & 101.22 \\

Tree + Summary
& 15.75 & 34.00
& 25.98 & \textbf{28.58}
& \textbf{19.69} & 66.83 \\

Tree + Selection + Summary
& 18.90 & 47.33
& 29.13 & 31.55
& 17.32 & 77.05 \\

\textbf{Full WMT}
& \textbf{33.86} & \textbf{32.48}
& \textbf{33.07} & 37.85
& 17.32 & \textbf{51.17} \\

\bottomrule
\end{tabular}
\caption{
Component-ablation results on GAIA-Text. Accuracy is the percentage of
correctly completed tasks, and prompt-token usage is reported in millions.
Bold values indicate the best result for each model and metric among the
reported tree-based configurations.
}
\label{tab:gaia-text-components}
\end{table*}

\begin{table*}[t]
\centering
\small
\setlength{\tabcolsep}{5pt}
\begin{tabular}{lcccccc}
\toprule
\textbf{Memory Configuration}
& \multicolumn{2}{c}{\textbf{Qwen3-8B}}
& \multicolumn{2}{c}{\textbf{Gemma 4 E4B}}
& \multicolumn{2}{c}{\textbf{Llama-3.1-8B}} \\
\cmidrule(lr){2-3}
\cmidrule(lr){4-5}
\cmidrule(lr){6-7}
& \textbf{Acc. (\%) $\uparrow$}
& \textbf{Tokens (M) $\downarrow$}
& \textbf{Acc. (\%) $\uparrow$}
& \textbf{Tokens (M) $\downarrow$}
& \textbf{Acc. (\%) $\uparrow$}
& \textbf{Tokens (M) $\downarrow$} \\
\midrule

Tree Memory
& 18.18 & 58.46
& 25.45 & 65.09
& 11.52 & 101.91 \\

Tree + Selection
& 21.21 & 100.51
& 26.67 & 105.57
& 16.36 & 134.15 \\

Tree + Summary
& 15.15 & 49.83
& 24.85 & \textbf{36.02}
& 16.97 & 94.25 \\

Tree + Selection + Summary
& 15.76 & 66.84
& 28.48 & 42.68
& 20.00 & 112.44 \\

\textbf{Full WMT}
& \textbf{29.70} & \textbf{42.69}
& \textbf{30.91} & 46.64
& \textbf{24.85} & \textbf{69.45} \\

\bottomrule
\end{tabular}
\caption{
Component-ablation results on GAIA. Accuracy is the percentage of
correctly completed tasks, and prompt-token usage is reported in millions.
Bold values indicate the best result for each model and metric among the
reported tree-based configurations.
}
\label{tab:gaia-full-components}
\end{table*}

\paragraph{Model-specific trends.}

The component ablations reveal different patterns across model families.
Qwen3-8B obtains its strongest partial-configuration accuracy from
Tree + Selection on both GAIA-Text and GAIA, suggesting that it benefits from
preserving fine-grained evidence and retrieving it selectively. Gemma 4 E4B
performs best among the partial configurations with
Tree + Selection + Summary, while Tree + Summary produces its lowest
prompt-token usage. This pattern suggests that summarization removes
redundant execution history for Gemma, while semantic selection remains
useful for recovering task-specific evidence.

Llama-3.1-8B exhibits a different trend. On GAIA-Text, Tree + Summary reaches
19.69\% accuracy, exceeding Full WMT at 17.32\%. On GAIA, however,
Tree + Selection + Summary is the strongest partial configuration, and Full
WMT achieves the best overall Llama result at 24.85\%. One plausible
explanation is that summary-only contexts sufficiently remove noisy
intermediate traces for text-only tasks, whereas the more heterogeneous GAIA
tasks benefit from combining compression with targeted, utility-aware
retrieval. These explanations are post hoc hypotheses rather than direct
measurements of model behavior.

\subsection{Memory-Poisoning Ablation Details}
\label{app:poisoning-ablations}

\paragraph{Evaluation corpus.}

The memory-poisoning evaluation contains 100 long-horizon scenarios composed
of 297 subtasks and 1,118 memory entries. Of these entries, 709 are benign and
409 are intentionally poisoned. Poisoned entries introduce misleading
observations, fabricated claims, or compromised tool-derived information into
the execution history so that they may influence later retrieval and
reasoning.

The same scenario structures and injected memories are used for each memory
configuration. The evaluation therefore isolates how memory organization,
retention scoring, lifecycle management, and semantic retrieval affect the
persistence and propagation of poisoned information.

\paragraph{Ablation configurations.}

We compare five memory configurations:

\begin{itemize}
    \item \textbf{Linear Memory} stores all task and action history in a
    single sequential context. No hierarchy, utility scoring, suppression, or
    selective retrieval is used.

    \item \textbf{Unweighted Tree (A1)} organizes memories into task and
    action branches but does not use retention scoring or branch
    prioritization.

    \item \textbf{No Memory Controller (A2)} computes retention scores and
    branch priorities but disables lifecycle operations, including folding,
    score-based suppression, and reopening.

    \item \textbf{No Semantic Retrieval (A3)} retains the hierarchical tree,
    retention scores, and Memory Controller, but prompt construction includes
    the full selected branch rather than semantically selecting individual
    memories.

    \item \textbf{Full WMT} uses hierarchical memory, dynamic retention
    scoring, branch-level priorities, lifecycle control, and semantic prompt
    construction.
\end{itemize}

Attack Success Rate, Poison Retrieval Rate, Infection Persistence, Blast
Radius, Amplification Factor, Context Compression Ratio, and Task Success
Rate are defined in Appendix~\ref{app:metric-equations}. We additionally
report average prompt-token usage and latency. The complete results are
reported in Table~\ref{tab:poisoning-results}.

\begin{table*}[t]
\centering
\scriptsize
\setlength{\tabcolsep}{5pt}
\begin{tabular}{lccccc}
\toprule
\textbf{Metric}
& \shortstack{\textbf{Linear}\\\textbf{Memory}}
& \shortstack{\textbf{Unweighted}\\\textbf{Tree (A1)}}
& \shortstack{\textbf{No Memory}\\\textbf{Controller (A2)}}
& \shortstack{\textbf{No Semantic}\\\textbf{Retrieval (A3)}}
& \shortstack{\textbf{Full}\\\textbf{WMT}} \\
\midrule

Attack Success Rate $\downarrow$
& 0.995 & 0.631 & 0.601 & 0.680 & \textbf{0.419} \\

Poison Retrieval Rate $\downarrow$
& 0.246 & 0.125 & 0.158 & 0.139 & \textbf{0.097} \\

Infection Persistence $\downarrow$
& 1.000 & \textbf{0.009} & 1.000 & \textbf{0.009} & \textbf{0.009} \\

Blast Radius $\downarrow$
& 0.906 & 0.388 & 0.509 & 0.669 & \textbf{0.315} \\

Amplification Factor $\downarrow$
& 6.103 & 1.280 & 1.638 & 1.977 & \textbf{0.965} \\

Task Success Rate $\uparrow$
& 0.183 & 0.431 & 0.451 & 0.393 & \textbf{0.575} \\

Context Compression Ratio $\downarrow$
& 1.000 & 0.815 & \textbf{0.634} & 1.000 & 0.819 \\

Average Prompt Tokens $\downarrow$
& 362.143 & 133.733 & 146.721 & 189.770 & \textbf{131.963} \\

Latency (ms) $\downarrow$
& 0.022 & 0.081 & 0.108 & \textbf{0.020} & 0.080 \\

\bottomrule
\end{tabular}
\caption{
Memory-poisoning results for WMT and its ablations. Arrows indicate the
preferred direction. Bold values indicate the best result for each metric;
ties are bolded jointly. Linear Memory retains the full sequential history,
A1 removes weighting, A2 removes lifecycle control, and A3 removes semantic
memory selection.
}
\label{tab:poisoning-results}
\end{table*}

\subsection{Worked Example of Memory-Poisoning Dynamics}
\label{app:poisoning-example}

Figure~\ref{fig:memory-poisoning-scenario} illustrates how WMT processes
ordinary and poisoned memories within a single execution tree. Blue nodes
represent the query and task hierarchy, gray nodes represent ordinary action
memories, and red nodes represent intentionally poisoned memories. The
example contains two task branches: identifying a candidate paper and
extracting its authors. Its purpose is to illustrate memory-state updates
rather than establish a canonical ranking of papers published in 2017.

\paragraph{Task initialization and event scores.}

WMT first creates a root node for the user query and an active child task for
identifying the paper. Each action performed under this task is inserted as a
child memory together with its observation and recorded outcome. As defined
in Eq.~\ref{eq:event-update}, the outcome assigns the action memory an
intermediate retention score before any selection feedback is applied. The
reported configuration uses

\[
(u_{\mathrm{success}},u_{\mathrm{failure}})
=
(0.75,0.30).
\]

Thus, a successful action receives
$\tilde{u}_i=0.75$, whereas a failed action receives
$\tilde{u}_i=0.30$. The tilde denotes the event-assigned intermediate
score. Selection-based decay may subsequently reduce this value, producing
the final retention score $u_i$.

\paragraph{Selection-based decay in the example.}

All action memories in Figure~\ref{fig:memory-poisoning-scenario} are
non-global memories. Let $u_i^{[r]}$ denote the score of memory $v_i$ after
$r$ consecutive missed selections, with
$u_i^{[0]}=\tilde{u}_i$. From
Eqs.~\ref{eq:decay-multiplier} and
\ref{eq:selection-decay}, the $r$th missed selection produces

\[
u_i^{[r]}
=
\operatorname{clip}_{[0,1]}
\left(
u_i^{[r-1]}
\rho^{\min\{r,M\}}
\right).
\]

Consequently, after $r$ consecutive missed selections,

\[
u_i^{[r]}
=
\operatorname{clip}_{[0,1]}
\left(
\tilde{u}_i
\rho^{h_M(r)}
\right),
\]

where

\begin{equation}
\begin{aligned}
h_M(r)
&=
\sum_{k=1}^{r}\min\{k,M\}
\\
&=
\begin{cases}
\dfrac{r(r+1)}{2},
& r\leq M,
\\[5pt]
Mr-\dfrac{M(M-1)}{2},
& r>M.
\end{cases}
\end{aligned}
\label{eq:cumulative-decay}
\end{equation}

This cumulative expression assumes that no intervening selection or new
event-based update occurs. Selecting the memory resets its missed-selection
count to zero, while a memory that does not enter the candidate pool is not
decayed.

\paragraph{First branch: identifying the paper.}

The first action searches broad encyclopedia pages. Although the returned
information does not provide a defensible ranking, the action is recorded as
operationally successful and therefore receives

\[
\tilde{u}_{\mathrm{encyclopedia}}=0.75.
\]

The second action searches citation summaries and produces evidence
supporting \emph{Attention Is All You Need}. It is likewise recorded as
successful:

\[
\tilde{u}_{\mathrm{citation}}=0.75.
\]

The first poisoned memory is introduced by an action that consults an
unofficial ranking blog. The observation contains an unsupported claim and
the action is explicitly recorded as a failure. Its event-assigned score is

\[
\tilde{u}_{\mathrm{blog}}=0.30.
\]

WMT preserves this failed action as a warning, but the lower score reduces
its priority as supporting evidence. The following action retrieves the
official arXiv record and is recorded as successful:

\[
\tilde{u}_{\mathrm{arXiv}}=0.75.
\]

Suppose the official arXiv record is selected during subsequent prompt
construction. Its missed-selection count is then reset to zero. If the
poisoned blog memory remains eligible but is not selected for $r$
consecutive opportunities, its score becomes

\[
u_{\mathrm{blog}}^{[r]}
=
\operatorname{clip}_{[0,1]}
\left(
0.30\,\rho^{h_M(r)}
\right).
\]

The poisoned record therefore remains in persistent memory, but its ability
to influence later prompts decreases with repeated missed selections. If it
does not enter the candidate pool, its score remains unchanged.

Once the paper-identification task is completed, the Memory Controller folds
the branch into a compact summary. The summary can preserve the supported
paper identity, the official source, and the warning about the failed blog
lookup without replaying the complete branch in subsequent prompts.

\paragraph{Second branch: extracting the authors.}

After the paper-identification branch is completed, WMT activates the task
for extracting the paper's authors. The first action opens the arXiv abstract
page. It verifies the title and identifier but does not recover the complete
author list. Because the action is operationally successful, it receives

\[
\tilde{u}_{\mathrm{abstract}}=0.75.
\]

The second poisoned memory is more difficult. An unofficial mirror page
returns a fabricated single-author claim, but the corresponding action is
recorded as successful. Event-based scoring therefore assigns

\[
\tilde{u}_{\mathrm{mirror}}=0.75,
\]

the same intermediate score assigned to a clean successful action. This case
demonstrates that the event outcome represents operational success rather
than factual correctness. A plausible but incorrect observation may therefore
receive a high initial score when the action itself appears successful.

The final action retrieves the official PDF and extracts the complete author
list. This action is recorded as successful and receives

\[
\tilde{u}_{\mathrm{PDF}}=0.75.
\]

Once the official result becomes available, the Prompt Synthesizer can select
it in preference to the mirror-page observation. If the mirror memory remains
eligible but is unselected for $r$ consecutive opportunities, its score
becomes

\[
u_{\mathrm{mirror}}^{[r]}
=
\operatorname{clip}_{[0,1]}
\left(
0.75\,\rho^{h_M(r)}
\right).
\]

Unlike the failed blog lookup, the mirror memory does not receive an immediate
low event score. Its influence must instead be reduced through semantic
selection, repeated missed-selection decay, or an explicit supersession rule.
If the official PDF result supersedes the mirror claim, the controller may
mark the poisoned memory \state{obsolete}. The memory remains stored for
provenance but is excluded from ordinary context construction.

\paragraph{Branch-priority updates.}

After the node scores are updated, WMT recomputes each task branch's priority
using Eq.~\ref{eq:branch-priority}. In the first branch, the failed blog
action increases the failed-action proportion $F_b$, thereby lowering the
branch priority. If a poisoned memory is subsequently marked
\state{obsolete}, it also contributes to the obsolete-node proportion
$O_b$.

The second branch initially contains no explicitly failed poisoned action
because the mirror lookup is recorded as successful. Consequently, the
failure-ratio penalty alone cannot distinguish the poisoned mirror memory
from a clean successful action. Its influence is instead controlled through
semantic selection, selection-based decay, and supersession by the official
PDF result. This illustrates the complementary roles of event-based score
assignment and selection-based feedback.

If the recomputed priority of an eligible branch satisfies

\[
U_b^{(t)}<\tau_{\mathrm{obs}}=0.10,
\]

the Memory Controller applies
Eq.~\ref{eq:obsolescence-transition} and marks the branch's root task node
\state{obsolete}. The branch and its descendants are then excluded from
ordinary context construction without being deleted. A single low-score
memory does not necessarily suppress an otherwise useful branch because
$U_b^{(t)}$ also incorporates the maximum node score and the mean utility of
the branch.

\paragraph{Final context construction.}

For the final reasoning step, the Prompt Synthesizer retains the active task
path and can include the folded paper-identification summary, the official
arXiv record, and the author list extracted from the official PDF. A compact
failure warning may also be included when it prevents the agent from
revisiting an untrusted source.

The poisoned blog and mirror memories remain in persistent storage but need
not enter the working context. The resulting prompt therefore contains the
supported paper identity and author information rather than the complete
execution history. The example demonstrates two distinct cases: an explicitly
failed poisoned memory begins with the lower event score
$u_{\mathrm{failure}}$, whereas a plausible poisoned memory recorded as
successful begins with $u_{\mathrm{success}}$ and must be controlled through
selection feedback, supersession, and lifecycle management.
\subsection{Inference Runtime and GPU Compute Usage}
\label{app:runtime-compute}

All experiments were inference-only and were executed on a server equipped
with eight NVIDIA Quadro RTX 6000 GPUs, each with 24\,GB of device memory,
for 192\,GB of aggregate installed GPU memory. Qwen3-8B, Gemma 4 E4B, and
Llama-3.1-8B were used as frozen reasoning models; no parameter training or
fine-tuning was performed. Independent benchmark jobs were scheduled across
the available GPUs.

The complete benchmark grid contains seven unique memory configurations,
three base models, and two evaluation sets, yielding 42
model--dataset--configuration settings. GAIA-Text contains 127 questions,
whereas GAIA contains 165 questions. Because each setting was evaluated
independently on its corresponding question set, the benchmark grid comprises

\[
R
=
7 \times 3 \times (127+165)
=
6{,}132
\]

question-level agent executions. Each execution initializes a new
query-specific memory tree. This count covers the GAIA-Text and GAIA
benchmark and component-ablation runs; the separate memory-poisoning
evaluation is not included.

The retained benchmark artifacts provide prompt-token totals but do not
contain a consistent per-execution wall-clock ledger, device-utilization
traces, or power measurements. We therefore do not report average runtime,
total wall-clock time, GPU-hours, floating-point operations, or energy
consumption. Prompt-token volume is instead used as the reproducible measure
of inference workload.

After deduplicating configurations that appear in both the main and component
tables, the seven unique configurations processed approximately 1.028 billion
prompt tokens on GAIA-Text and 1.363 billion on GAIA, for 2.391 billion prompt
tokens across the complete benchmark grid. Full WMT accounts for
121.50 million prompt tokens on GAIA-Text and 158.78 million on GAIA, or
280.28 million in total.

These totals include all language-model calls required by each configuration,
including base-agent reasoning, semantic memory selection, and branch
summarization when enabled. Because the three model families use different
tokenizers, the combined token count represents aggregate inference workload
rather than a tokenizer-normalized comparison across models.

The reported 192\,GB denotes aggregate installed memory across the eight
GPUs and should not be interpreted as the memory consumed by an individual
run. Peak allocated GPU memory was not recorded.

\end{document}